\documentclass{article}
\usepackage{ijcai21}

\usepackage{times}

\usepackage{soul}
\usepackage{url}
\usepackage[hidelinks]{hyperref}
\usepackage[utf8]{inputenc}
\usepackage[small]{caption}
\usepackage{graphicx}
\usepackage{amsmath}
\usepackage{booktabs}
\usepackage{xcolor}
\usepackage{amsfonts}
\usepackage{multirow}
\usepackage{algorithm}
\usepackage{algorithmicx}
\usepackage{algpseudocode}
\usepackage{amsmath}
\usepackage{verbatim}
\usepackage{graphicx}
\usepackage[switch]{lineno}

\title{
	Novelty Detection via Contrastive Learning with Negative Data Augmentation}

\newcommand\correspondingauthor{\thanks{Corresponding authors.}}

\author{
	Chengwei Chen$^1$\thanks{These authors contributed equally to this work.}\and
	Yuan Xie$^1$$^*$  \and
	Shaohui Lin$^1$$^\dagger$ \and
	Ruizhi Qiao$^2$ \and
	Jian Zhou$^2$ \and 
	Xin Tan$^3$ \and
	\\
	Yi Zhang$^4$ and
	Lizhuang Ma$^1$\correspondingauthor \ \\
	\affiliations
	$^1$East China Normal University\\
	$^2$Tencent Youtu Lab\\
	$^3$Shanghai Jiao Tong University\\
	$^4$Zhejiang Lab\\
	\emails
	52184501028@stu.ecnu.edu.cn, 
	\{yxie,shlin\}@cs.ecnu.edu.cn,
	 \{ruizhiqiao,darnellzhou\}@tencent.com, tanxin2017@sjtu.edu.cn, zhangyi620@zhejianglab.com, lzma@cs.ecnu.edu.cn
}

\begin{document}
	
\maketitle

	\begin{abstract}
Novelty detection is the process of determining whether a query example differs from the learned training distribution. 
Previous generative adversarial networks based methods and self-supervised approaches suffer from instability training, mode
dropping, and low discriminative ability. We overcome such problems by introducing a novel decoder-encoder framework. Firstly,  a generative network (\emph{a.k.a.} decoder) learns the representation by mapping the initialized latent vector to an image. In particular, this vector is initialized by considering the entire distribution of training data to avoid the problem of mode-dropping. Secondly, a contrastive network (\emph{a.k.a.} encoder) aims to ``learn to compare'' through mutual information estimation, which directly helps the generative network to obtain a more discriminative representation by using a negative data augmentation strategy. Extensive experiments show that our model has significant superiority over cutting-edge novelty detectors and achieves new state-of-the-art results on various novelty detection benchmarks, \emph{e.g.} CIFAR10 and DCASE. Moreover, our model is more stable for training in a non-adversarial manner, compared to other adversarial based novelty detection methods.
		\end{abstract}

	\section{Introduction}
Novelty detection can be described as a one-class classification, which aims to detect the samples whether drawing far away from the learned distribution of training samples from the target class. Generative adversarial networks (GANs) \cite{sabokrou2018adversarially,perera2019ocgan,ijcai2020-106} have been a common choice for novelty detection. Generator and discriminator compete mutually while collaborating to learn a representative latent space for the target class. In this latent space, the reconstructed features of novelty samples (outliers) have higher reconstruction errors than normal samples (inliers), which can be used for distinguishing between normal and novelty classes. Recently, self-supervised learning \cite{komodakis2018unsupervised,ji2019invariant} holds great for improving representations when labeled data are scarce. In the training process, the network learns useful feature representation by solving some specialized pretext tasks, such as geometric transformations prediction \cite{hendrycks2019using}. During inference, the model is transferred to the downstream task, like novelty detection. 
	
	\begin{figure}[t!]
		\centering
		\includegraphics[width=\linewidth]{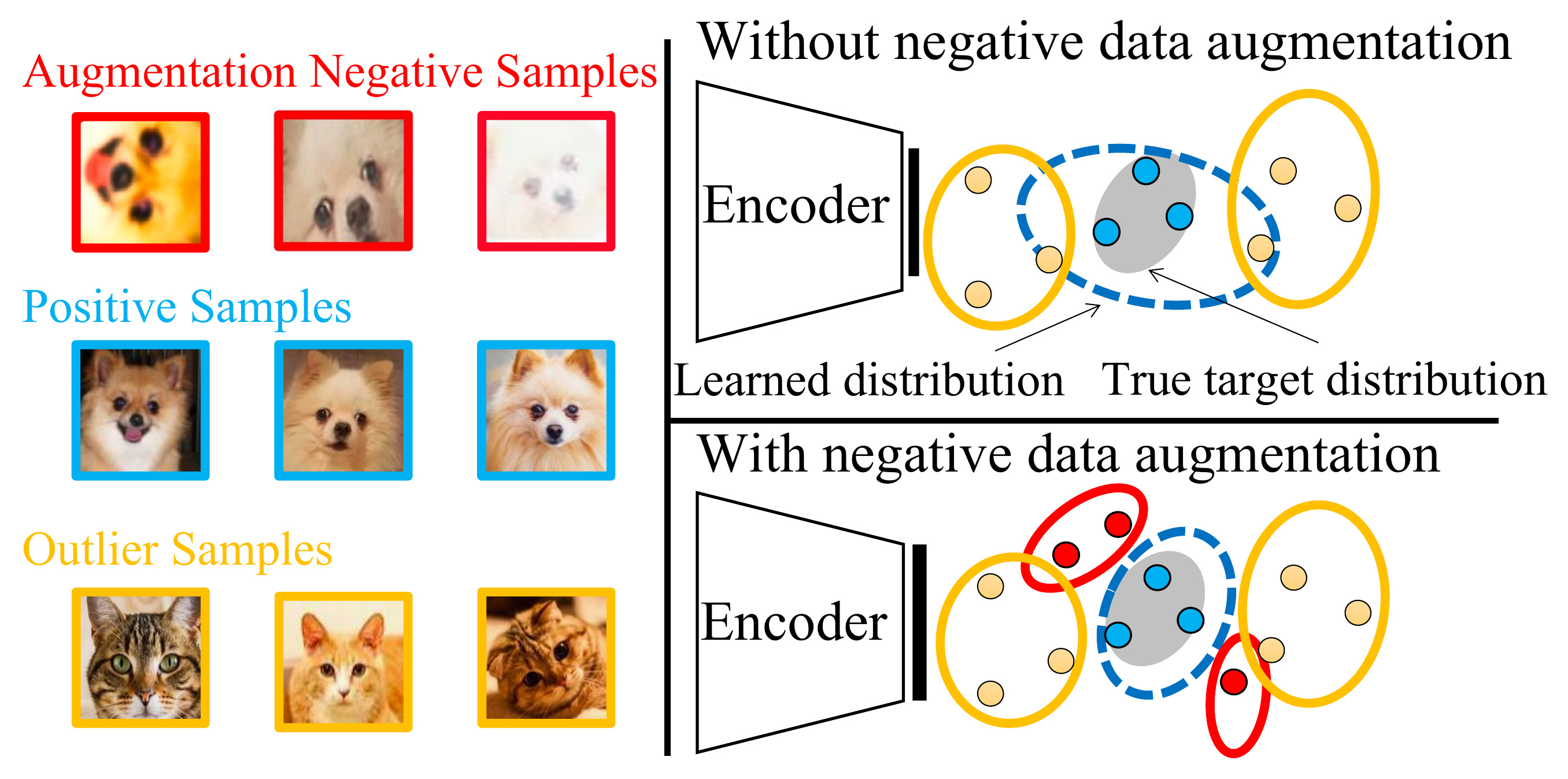}
		\caption{
			Overview of negative data augmentation strategy. \textbf{Top}: without the negative data augmentation strategy, the learned distribution (blue oval) of target samples may include the outliers (yellow dots). \textbf{Bottom}: The novelty-like samples (red dots) are generated from target samples by negative data augmentation where we employ multiple transformations combination. Compared with outliers, the distribution of novelty-like samples (red oval) is more close to that of the target samples. Through mutual information estimation, the learned distribution of target class can be pulled to close to the true target distribution (gray oval), and pushed to far away from the distribution of hard novelty-like samples.
		}
		\label{vs}
	\end{figure}

However, some weaknesses are still existing in the previous work. For the GAN based methods, it suffers from three critical problems: mode-dropping, instable training, and low discriminative ability. First, various GAN based methods (\emph{e.g.} ALOCC \cite{sabokrou2018adversarially}, OCGAN \cite{perera2019ocgan} and DualGAN \cite{ijcai2020-106}), only learn the partial modes of target distributions, which causes the problem of mode dropping \cite{arora2017gans}. Second, the imbalance capacity of generator and discriminator causes the training of model unstable \cite{zhao2016energy}, which affects the learning of latent representation of normal samples. Third, the latent features with low discriminative ability \cite{liu2020self} are generated in self-representation, since the decoder of GANs tends to learn more structive representation than discriminative characteristics. 

The existing self-supervised learning based novelty detection requires specialized implementation for the pretext tasks, such as design supervised labels, loss functions, and network architectures. Rotation prediction  \cite{komodakis2018unsupervised} and transformations prediction \cite{hendrycks2019using} could capture the semantic information of object shapes that are useful for target tasks. However, it lacks other semantic information such as object textures and colors, which leads to the low discriminative ability of features that fails to effectively detect novelty samples. Besides, some \cite{lim2018doping,sinha2021negative} use data augmentation as an additional source of data in the GAN. Inspired by these works,  contrasting shifted instance (CSI) \cite{tack2020csi} contrasts distributionally-shifted augmentations with an auxiliary softmax classifier to learn the feature representation of encoder based on SimCLR \cite{chen2020simple}. However, the augmented images as negative samples only use one random augmentation, which cannot generate effective negative samples. Beside, the detection score function of CSI is complex with high computation and memory cost.

\begin{figure*}[h]
		\centering
		\includegraphics[width=\linewidth]{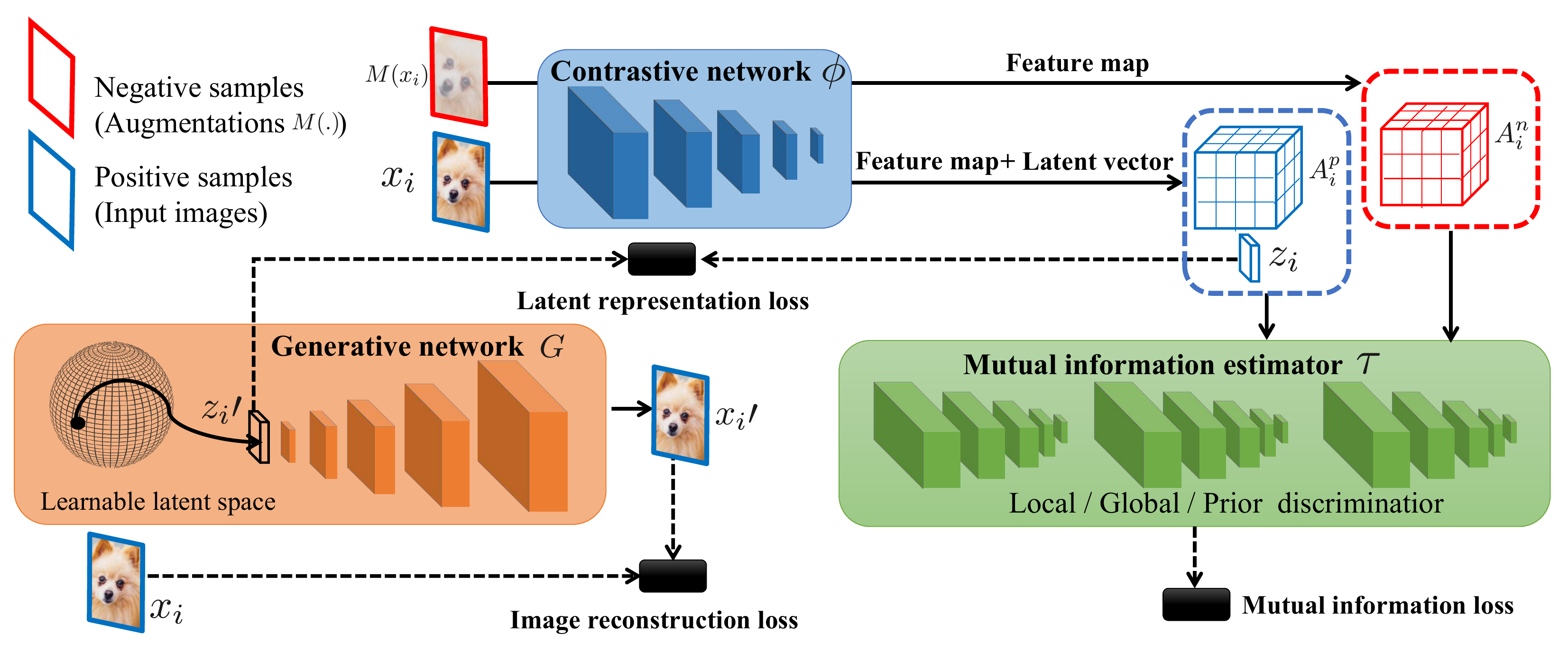}
		\caption{Illustration of our decoder-encoder framework for novelty detection. It consists of three components: a generative network, a contrastive network, and a mutual information estimator. The generative network (decoder) learns a reconstructed image representation by mapping each initialized latent vector to each target image. The contrastive network (encoder) encodes a pair of target and novelty-like samples from data augmentation to extract their latent global features and local feature maps. The mutual information estimator is adopted to generate discriminative latent features through contrastive learning on the features of input pairs by maximizing the local, global and prior mutual information.}
		\label{architecture}
	\end{figure*}

To address these issues, we propose a novel decoder-encoder framework for one-class novelty detection to learn more discriminative latent representation by \textit{contrastive learning}. Our framework consists of three parts: a generative network, a contrastive network, and a mutual information estimator. First, the generative network (decoder) aims to learn the representation of  target class by mapping each initialized latent vector to each target image; The initialization of the latent vectors is obtained by encoding the entire distribution of training data to alleviate the problem of mode-dropping. Then, we employ the contrastive network (encoder) to extract both local feature maps from positive and negative samples. It also captures the global latent features from positive samples. In particular, we select the normal training data as positive samples and use negative data augmentation strategy to generate hard negative samples (\emph{a.k.a.} novelty-like samples) from their corresponding positive samples by using multiple random transformations. Different from outliers, these novelty-like samples are more close to the normal ones. Therefore,  it can better help to separate the distribution of normal and novelty samples by learning the disjointness between positive and novelty-like samples, as illustrated in Fig.\ref{vs}. Finally, mutual information estimator is adopted to generate discriminative latent features through contrastive learning on the features of input pairs by maximizing the local, global, and prior mutual information. Our decoder-encoder framework is presented in Fig. \ref{architecture}.

	We summarize  our main contributions of this paper:
	
	\begin{itemize}
		
\item We propose a novel effective decoder-encoder framework for the novelty detection task. Contrastive learning is introduced to learn more discriminative latent features for distinguishing between positive and negative augmentation samples.

\item The mutual information estimator is trained in a non-adversarial way by distinguishing between features of local parts and global context constituted by only positive and novelty-like samples, which helps the model to train more stably with faster convergence than GAN-based methods.

\item  Extensive experiments demonstrate the superior performance of our approach for novelty detection in several challenging datasets. For instance, our method achieves the highest mean AUC of 0.843 and 0.899 on CIFAR-10 and DCASE, compared to state-of-the-art methods.

	\end{itemize}

	\section{Methodology}

	\subsection{Our Decoder-encoder Framework}
	 Our framework consists of three components: a generative network, a contrastive network and a mutual information estimator, as it shown in Fig. \ref{architecture}. The generative network learns a mapping from  latent space to  high-dimensional image space, while the contrastive network maps a positive/negative image to local feature maps and a global latent feature. The mutual information estimator is used for distinguishing between the features from the target samples and their corresponding hard negative samples to effectively learn more discriminative latent feature presentation.
	
	\paragraph{Notations.}
	Let $X = \{x_1,\cdots, x_N\}$ denote the original input images of target class with $N$ samples, $Z = \{z_1,\cdots, z_N\}$ denote their corresponding global latent features where $z_i=\phi_{\theta_c}(x_i)\in\mathbb{R}^d$ is learned by the contrastive network $\mathbb{C}$ with parameters $\theta_c$. $d$ is the dimension of latent features. 
	$A_i=\phi_{\theta_{fp}}(x_i)\in\mathbb{R}^{H\times W\times S}$ is the local feature maps extracted from contrastive network $\mathbb{C}$, where $\theta_{fp}\subset\theta_c$ and $H\times W\times S$ is the dimension of feature maps.
	$ x^{\prime}_i =  \psi_{\theta_g}(z^{\prime}_i)$ presents the reconstructed image by generative network $\mathbb{G}$ with parameters $\theta_g$, where $z^{\prime}_i$ is the initialized latent vector. In mutual information estimator, we denote the global estimator, local estimator and prior estimator as $\mathbb{GE}, \mathbb{LE}$ and $\mathbb{PE}$ with parameters $\theta_{ge},  \theta_{le}$ and $\theta_{pe}$, respectively.
	
	\paragraph{Generative network.}
	The input latent vectors ${z^{\prime}_1,,\cdots, z^{\prime}_N}$ of generative network are first initialized by the PCA projection of all input images, which helps to alleviate the problem of mode dropping. Like a decoder, the generative network outputs $\psi_{\theta_{g}}(  z^{\prime}_i  )$  should be regained close to the input image $x_i$. Instead of MSE loss, we employ the Laplacian pyramid loss \cite{ling2006diffusion} as the reconstruction loss; MSE is easy to yield the blurry image, while the Laplacian pyramid loss can generate better reconstructed images by Laplacian pyramid representation. Therefore, the reconstruction loss between the reconstructed output $\psi_{\theta_{g}}(z^{\prime}_i)$  and the input image $x_i$ is formulated as:

	\begin{equation}
	\footnotesize
	L_{lap}=\sum_{j} 2^{2 j}\left|Lap^{j}(\psi_{\theta_{g}}(z^{\prime}_i)) - Lap^{j}\left(x_{i}\right)\right|_{1},
	\label{lap}
	\end{equation}
	where  $Lap^j(\cdot)$ is the $j$-th level of Laplacian pyramid representation. 
	
	\paragraph{Contrastive network.}
	To improve the discriminative ability of the latent features of the target class, the contrastive network uses a pair of positive and negative samples as an input to extract both local feature maps $(A_i^p, A_i^n)$  and a global latent vector $z_i$ of the positive sample, where the samples from different classes compete each other through mutual information estimator. As shown in Fig. \ref{architecture} (blue box), a positive samples $x_i$ is an original image and its negative sample is generated by the negative data augmentation $M$ as $M(x_i)$ (called novelty-like samples). Our transformations of augmentation include random resized crop, random color jitter, random grayscale, and random horizontal flip. We employ the combination of multiple random transformations to capture more rich semantic information, including object shape, textures, and colors. In addition, this augmentation can better help to separate the distribution of normal and novelty samples by learning the disjointness between positive and novelty-like samples, as the distribution of novelty-like samples is more close to positive ones than outliers. In our framework, we employ cooperative learning between the generative network and contrastive network to generate better reconstructed images and provide more discriminative global latent vectors. It also motivates us to use the decoder-encoder framework instead of the conventional encoder-decoder frameworks with unilateral learning \cite{vincent2010stacked,marchi2017deep}. Therefore, we need to minimize the distance between the initialized latent vector $z^{\prime}_i $ and the global latent vector $\phi_{\theta_c}(x_i)$ as: 

	\begin{equation}
	\footnotesize
	L_{lat}= \| z^{\prime}_i -	\phi_{\theta_c}(x_i)\|_2^2.
	\label{aux}
	\end{equation} 
	
	Inspired by \cite{sabokrou2018adversarially}, the reconstruction space is more effective for distinguishing the positive and novelty samples compared to the latent space. Therefore, a test example $x$ first go through contrastive network and then generative network in a encoder-decoder manner during testing. The constraint by Eq. (\ref{aux}) makes the testing feasible. Without this constraint, the reconstructed images from positive samples will have significantly large reconstructed error even with the constraint of Eq. (\ref{lap}).

	\begin{figure*}[t]
		\centering
		\includegraphics[width=\linewidth]{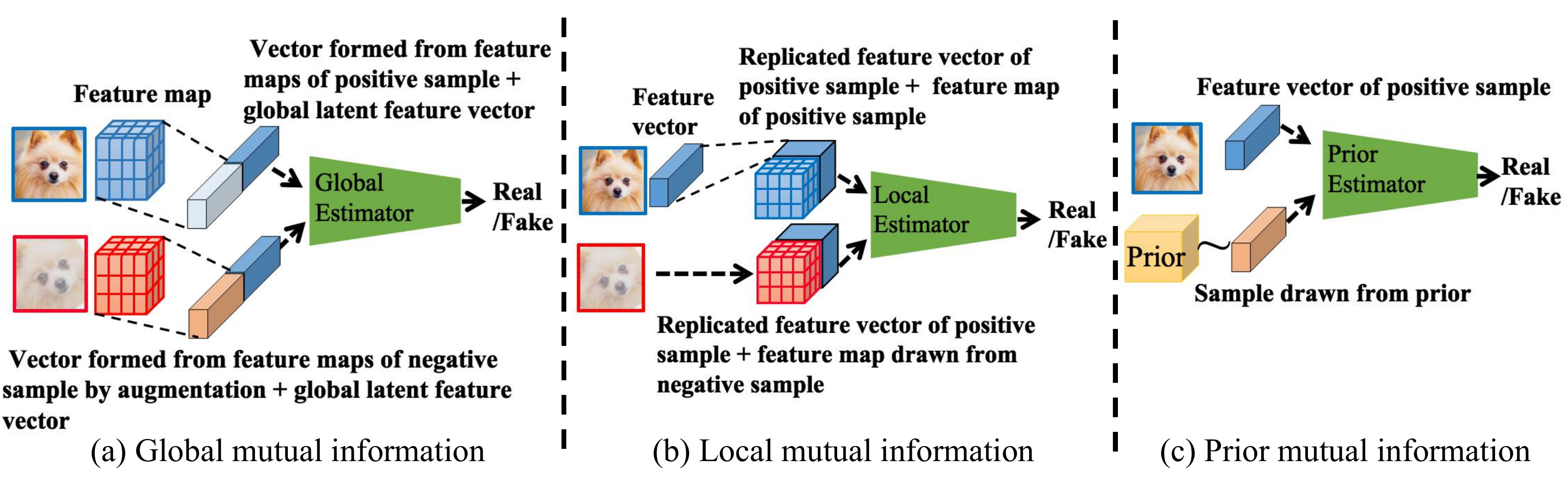}
		\caption{Local, global and prior mutual information estimation.}
		\label{estimation}
	\end{figure*}
	
	\paragraph{Mutual information estimator.}
	Mutual information measures the essential relevance of two instances. The larger mutual information is, the more similar two variables have. Given two random variables $x$ and $y$, mutual information \cite{belghazi2018mine} can be estimated by the JS divergence between the joint $p(y\mid x)p(x)$  and the product of the margins $p(x)p(y)$. According to the definition of the variational estimation of JS divergence \cite{nowozin2016f}, the maximization of the mutual information  between variables X and Y can be formulated as: 

	\begin{equation}
	\scriptsize
	\begin{aligned} \min _{\theta_{e}} -I(X, Y)=\min _{\theta_{e}}\left\{- \left (\mathbb{E}_{(x, y) \sim p(y \mid x) p(x)}[\log \sigma(T(x, y))]\right.\right.& \\\left.+\mathbb{E}_{(x, y) \sim p(y) p(x)}[\log (1-\sigma(T(x, y)))]\right) \}, \end{aligned}
	\label{mut5}
	\end{equation}
	where $\sigma$ denotes the sigmoid function and \(T(x)=\log \frac{2 p(x)}{p(x)+q(x)}\). Here $p(z|x)p(x)$ and $p(z)p(x)$ are utilized to replace $p(x)$ and $q(x)$.

	In our paper, we divide mutual information estimator into three parts: global estimator, local estimator and prior estimator. The goal of the mutual information estimator is to generate discriminative global latent features by contrastive learning between positive samples and negative samples. We first consider the global mutual information (see Fig. \ref{mut5}(a)).  Based on Eq. (\ref{mut5}), the maximization of global mutual information is also equivalent to minimize Eq. (\ref{mut6}) by introducing a global estimator $\tau_{\theta_{ge}}([A_z, z])$, which can be formulated as:

	\begin{equation}
	\scriptsize
	\begin{aligned} & L_{global}= - {\beta}  \left(\mathbb{E}_{(A^p, z) \sim p(z \mid A^p) p(A^p)}[\log \sigma(\tau_{\theta_{ge}}([A_z^p,z]))]\right.\\ &\left.\left.+\mathbb{E}_{(A^n, z) \sim p(z) p(A^n)}[\log (1-\sigma( \tau_{\theta_{ge}}([A_z^n,z])))]\right),  \right.\ \end{aligned}
	\label{mut6}
	\end{equation}
	where $A_z^p$ and $A_z^n$ are the vectors downscaled from the feature maps $A^p$ and $A^n$ from the positive sample and the negative sample, respectively. They all have the same dimension to $z$. $[A_z, z]$ is the concatenation between the downscaled feature $A_z$ and $z$ as an input pair. $\beta$ is a hyperparameter. Similar to \cite{hjelm2019learning}, the optimization of Eq. (\ref{mut6}) is to estimate the global latent feature distribution from positive samples by distinguishing the input from positive samples or negative samples.
	
	Second, we consider local mutual information (see Fig. \ref{estimation}(b)), and also construct the relationship between the local feature map and the global latent feature. The process of estimation is the same as global mutual information. Thus, the objective function of local mutual information loss can be formulated as:

	\begin{equation}
	\scriptsize
	\begin{array}{l} L_{local}= -\frac{\beta}{HW} \Sigma_{i, j}\left(\mathbb{E}_{(A^p, z) \sim p(z \mid A^p) p(A^p)}\left[\log \sigma\left(\tau_{\theta_{le}}([A_{ij}^p, z_A])\right)\right]\right. \\ \left.+\mathbb{E}_{(A^n, z) \sim p(z) p(A^n)}\left[\log \left(1-\sigma\left(\tau_{\theta_{le}}([A_{ij}^n,z_A])\right)\right)\right]\right),\end{array}
	\label{mut7}
	\end{equation}
	where  $\tau_{\theta_{le}}([A_{ij}, z_A])$ is a local estimator to output the reverent probability between the local feature map $A$ and latent representation $z_A$, which consists of a wide range of replicated feature vectors from $z$ with the same dimension to $A$.  An input pair is either from a positive sample $[A_{ij}^p, z_A]$ or a negative sample $[A_{ij}^n, z_A]$ at coordinates $(i,j)$. $H$ and $W$ represent the height and width of the feature map.

	Third, we employ the KL-divergence between the global latent feature and the prior distribution (\emph{e.g.}, normal distribution) to encourage the latent feature to be more regular. Thus we can construct the following objective function:

	\begin{equation}
	\footnotesize
	\begin{aligned}  & L_{prior}=\ {\gamma}  	\mathbb{E}_{A \sim p(A)}[K L(p(z \mid A) \| q(z))],  \end{aligned}
	\label{mut8}
	\end{equation}
	where $q(z)$ is a prior distribution (\emph{e.g.}, normal distribution) and $p(z|A)$ is the output of prior estimator $\tau_{\theta_{pe}}$.
	By combining the aforementioned three loss functions (\emph{i.e.}, Eqs (\ref{mut5}), (\ref{mut6}) and (\ref{mut7})), we obtain the final mutual information estimator loss function:

	\begin{equation}
	L_{mie} = L_{global} + L_{local} + L_{prior}.
	\label{mutual}
	\end{equation}
	By minimizing the above function, we generate discriminative latent features $z$ that helps to make the normal and novelty samples separable.

	\subsection{Overall Loss Function}
	According to the above discussion, we can construct the overall loss function for our decoder-encoder framework as:
	\begin{equation}
	L_{all} = \lambda_1 L_{lap} + \lambda_2 L_{lat} + \lambda_3 L_{mie}
	\end{equation}
	where $\lambda_1, \lambda_2$ and $\lambda_3$ are the hyperparameters for balancing these three different terms.
	For solver, SGD optimizer can be directly used to minimize Eq. (8) in an end-to-end manner.

	\subsection{Implementation Details } We adopt the structure of DCGAN \cite{radford2015unsupervised} as the structure of the generative network and contrastive network. The global mutual information estimator is a fully-connected network with
	two 512-unit hidden layers.  A 1 x 1 convnet with two 512-unit hidden layers is regarded as the local mutual information estimator.  The prior mutual information estimator is a fully-connected network with two hidden layers of 1000 and 200 units. 
	
	Our image augmentation $M$ contains cropping, horizontal flip, color jitter, rotation, and grayscale for random augmentations. During inference, the test sample $x$ first goes through the contrastive network to be encoded into a latent vector. The generative network then upscales the latent vector to reconstruct the image from the learned discriminative latent feature space. Finally, the abnormal score is calculated by image reconstruction error between test image sample $x$ and the corresponding generated image $x{'}$. The test sample $x$ is regarded as a novelty instance if the image reconstruction error is larger than a predefined threshold $T$, and a normal instance otherwise. We use PyTorch \cite{paszke2017automatic} to implement our method. For training parameters, the learning rate and the number of total epochs are set to 0.002 and 100, respectively. SGD optimizer with momentum is adopted to optimize the parameters of our framework. Batch size, momentum and weight decay are set to 128, 0.9 and 0.005, respectively. For hyperparameters, $\beta$ and $\gamma$ are set to 0.5 and 0.1, respectively. $\lambda_{1}$, $\lambda_{2}$ and $\lambda_{3}$ are all set to 1.

	\section{Experiments}
	
	\subsection{Experimental Setting}
	
		\paragraph{Public novelty detection dataset.}
	 We select CIFAR-10 \cite{krizhevsky2009cifar}, COIL 100 \cite{Nene1996}, MNIST \cite{1998Gradient}, fMNIST \cite{xiao2017/online} and DCASE \cite{DCASE2017challenge}  as the standard evaluation dataset. COIL100 contains 100 objects with multiple different poses, where each class has less than 100 images. MNIST contains 60K/10K 28 $\times$ 28 training/test gray-scale handwritten digit images from  0 to  9. fMNIST contains 60K/10K 28 $\times$ 28 training/test images of fashion apparels/accessories. CIFAR-10 contains 50K/10K 32 $\times$ 32 training/test images with diverse content, background and complexity from 10 classes. In DCASE \cite{DCASE2017challenge} dataset, all abnormal event audios are artificially mixed with background audios ({\it \emph{e.g.}} home, bus, and train) which contains 15 different acoustic background scenes.  We use the original mixed audio files	from the challenge, which contains 491, 496 and 500 audio files of roughly 30 seconds in the training, validation and test dataset, respectively. This task aims to distinguish abnormal acoustic signals from the normal ones. 
		
			\begin{table}[t!]
				\tiny
				\centering 
				\begin{tabular}{l|c|c|c|c|c|c|c|c}
					\hline 
					Local MI &&$\surd$&&&$\surd$&$\surd$&&$\surd$ \\	 
					Global MI &&&$\surd$&&&$\surd$&$\surd$&$\surd$ \\	
					Prior MI &&&&$\surd$&$\surd$&&$\surd$&$\surd$ \\	 \hline 
					CIFAR-10  &0.750&0.832&0.838&0.823&0.841&0.842&0.842&0.843 \\	 \hline 
				\end{tabular}
				\caption{The effect of different mutual information estimation.}
				\label{losscomposition}
			\end{table}

			\begin{figure}[t!]
				\centering
				\includegraphics[width=\linewidth]{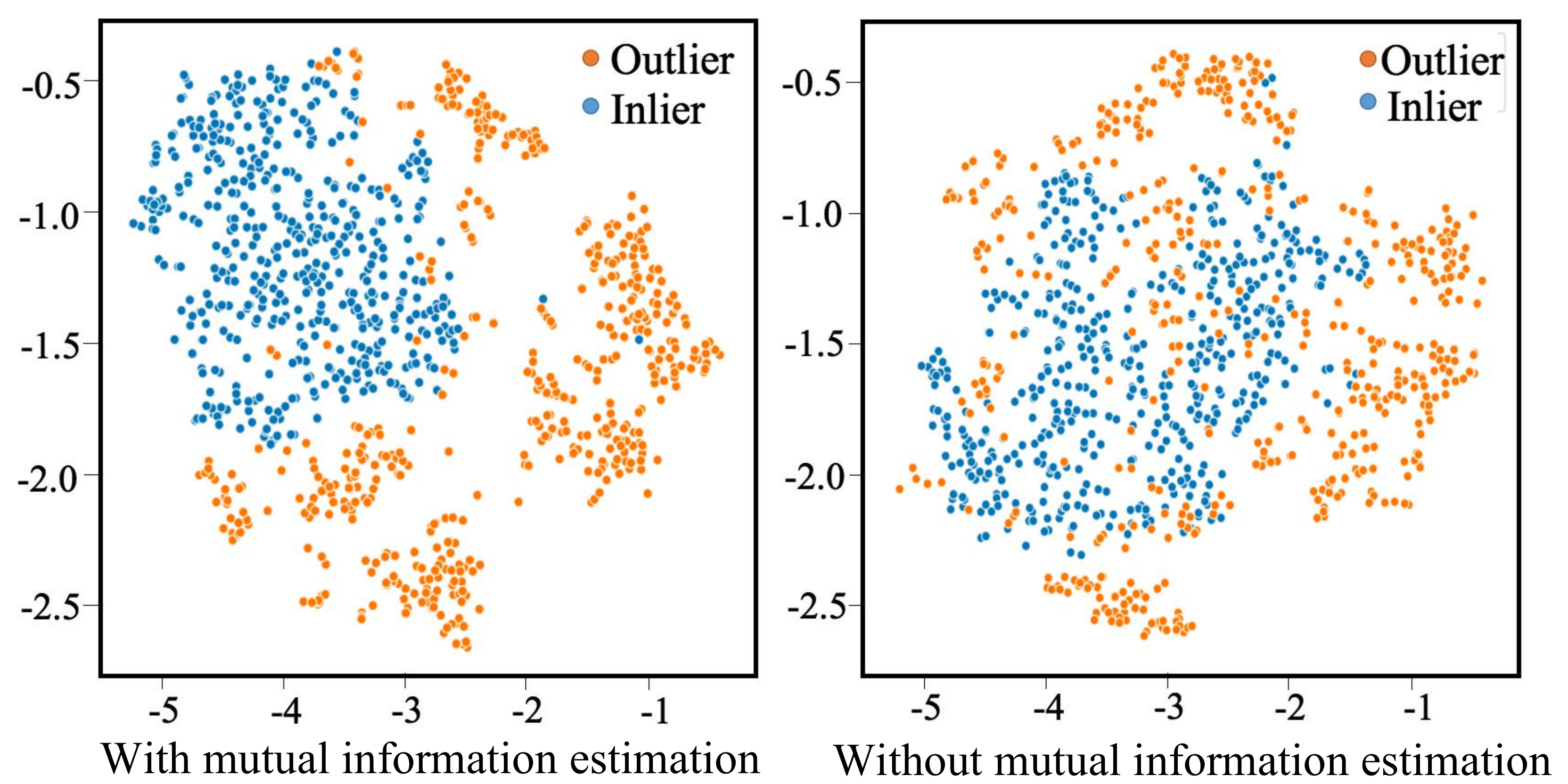}
				\caption{The visualization of latent space  learned from target class (Frog class) by using proposed method with/without mutual information estimation.}
				\label{latent}
			\end{figure}
			
			\begin{table}[t!]
			\tiny
				\centering 
				\begin{tabular}{l|c|c|c|c|c|c|c|c|c}
					\hline 
				\multirow{2}*{Dataset}&\multicolumn{3}{|c|}{Removal Feature} &\multicolumn{2}{|c|}{Reconstruction Loss}&\multicolumn{3}{|c|}{Different Prior}&	\multirow{2}*{AUC} \\
				\cline{2-9}
				  &$z_A$&z&None&MSE&Laplacian&N&C&P& \\	 \hline 
					\multirow{7}{*}{\quad \rotatebox{90}{CIFAR-10}} 
					&	$\surd$& 	&	&	&	$\surd$& & 	&$\surd$&0.828 \\	
					&	& $\surd$	&	&	&$\surd$	& && $\surd$	&0.821 \\	
					&	& 	&$\surd$	&	&$\surd$	& & 	&$\surd$&0.843  \\	
					&	& 	&$\surd$	&	&	& & 	&$\surd$&0.530  \\	 
					&	& 	&$\surd$	&	$\surd$&	& & &$\surd$	&0.810 \\	
					&	& 	&$\surd$	&	&$\surd$	&$\surd$ & 	&&0.745 \\	
					&	& 	&$\surd$	&	&$\surd$	& &$\surd$ 	&&0.795 \\
					\hline 
				\end{tabular}
				\caption{Ablation  study for proposed method. Part1:Removal of information inputs for the estimators. Part2:Comparsion of different reconstruction losses in generative network. Part3:The effect of different priors in generator inputs. N=Normal distribution, C=Contrastive prior, P=PCA prior.}
				\label{ab}
			\end{table}
		
			\paragraph{Face anti-spoofing detection dataset.}
		Replay-Attack \cite{chingovska2012effectiveness} dataset and CASIA-MFSD \cite{zhang2012face} dataset contain different attacks \emph{e.g.} printed paper face, replaying a video and wearing a mask. Replay-Attack dataset contains 1.2K videos (200 real access videos and 1K attack videos).  CASIA-MFSD  dataset contains 50 subjects where each subject has 12 videos under 3 different image resolutions and varied lightings. The goal of this task is to detect whether a face is “alive” or just a fraudulent reproduction. 
		
		\paragraph{Evaluation methodology. }
	The protocol in the literature is proposed for one-class novelty detection \cite{perera2019ocgan}. All of in-class training samples from only one class are used for training, and all samples in test set are used for testing. We use  Area Under Curve (AUC) to evaluate the performance in novelty detection and acoustic anomaly detection. We also use Half Total Error Rate (HTER) \cite{bengio2004statistical} for spoofing face detection.

\begin{table*}[t!]
	\tiny
	\centering 
	\renewcommand\tabcolsep{10.0pt} 
	\begin{tabular}{c||c|c|c|c|c|c|c|c|c|c||c}
		\hline 
		Methods&\textbf{PLANE} &\textbf{CAR}&\textbf{BIRD}&\textbf{CAT}&\textbf{DEER}&\textbf{DOG}&\textbf{FROG}&\textbf{HORSE}&\textbf{SHIP}&\textbf{TRUCK}&\textbf{Mean}\\
		\hline 
		\textbf{OCSVM  ('01)}&0.630 &0.440 &0.649 &0.487& 0.735& 0.500 &0.725& 0.533& 0.649& 0.508& 0.586\\
		
		\textbf{VAE  ('13)}&0.700&0.386&0.679&0.535&0.748&0.523&0.687&0.493&0.696&0.386&0.583\\
		
		\textbf{AnoGAN  ('17)}&0.671&0.547&0.529&0.545&0.651&0.603&0.585&0.625&0.758&0.665&0.618\\
		
		\textbf{DSVDD  ('18)}&0.617&0.659&0.508&0.591&0.609&0.657&0.677&0.673&0.759&0.731&0.648\\
		
		\textbf{ALOCC  ('18)}&
		0.620 &0.717 &0.537 &0.560 &0.587 &0.563 &0.612 &0.605& 0.744& 0.671 &0.622\\			
		
		\textbf{RotNet  ('18)}&
		0.719 &\textbf{0.945} &0.784 &0.700 &0.772 &\textbf{0.866} &0.816 &\textbf{0.937}& 0.907& \textbf{0.888} &0.833\\					
		
		\textbf{Neighbour  ('19)}&0.690&0.442&0.683&0.513&0.767&0.500&0.724&0.511&0.691&0.433&0.613\\
		
		\textbf{AND  ('19)}&0.717&0.494&0.662&0.527&0.736&0.504&0.726&0.560&0.680&0.566&0.617\\
		
		\textbf{IIC  ('19)}&0.684&0.894&0.498&0.653&0.605&0.591&0.493&0.748&0.818&0.767&0.674\\
		
		\textbf{Geometric  ('19)}&
		0.762 &0.848 &0.771 &0.732 &0.828 &0.848 &0.820 &0.887& 0.895& 0.834 &0.823\\		
		
		\textbf{OCGAN  ('19)}&0.757&0.531&0.640&0.620&0.723&0.620&0.723&0.575&0.820&0.554&0.657 \\

		\textbf{AE-EN  ('20)}&
		0.791&0.602&0.644&0.596&0.724&0.638&0.712&0.615&0.701&0.724&0.675 \\
		
		\textbf{DROCC  ('20)}&
		0.817&0.767& 0.667 &0.671 &0.736 &0.744 &0.744 &0.714 &0.800 &0.762& 0.742 \\	
		
		\textbf{DualGAN  ('20)}&
		0.875&0.548&0.719&0.639&0.833&0.643&0.810&0.581&0.872&0.503&0.703 \\
		\hline
		
		\textbf{Ours}&\textbf{0.985}	&0.765&\textbf{0.796}&\textbf{0.791}&\textbf{0.924}&0.717&	\textbf{0.975}&0.691&\textbf{0.985}&0.752&	\textbf{0.843}\\
		\hline 
	\end{tabular}
	\caption{AUC of different novelty detection methods on CIFAR-10. Plane and car  denote Airplane and Automobile in CIFAR-10, respectively.}
	\label{cifar10}
\end{table*}

	\subsection{Ablation Study}
In this part, we evaluate the effect of each mutual information loss in Eq. (\ref{mutual}), the effect of information fusion between feature maps and latent feature, the effect of Laplacian pyramid loss and the effect with/without PCA initialization. We conduct the experiments on CIFAR-10 for ablation study.

\paragraph{The effect of mutual information estimation.}
 We first evaluate the effect of each component in mutual information estimation. The results are summarized in Tab. \ref{losscomposition}. Obviously, our model achieves a significantly higher  AUC score by 0.843 with all mutual information estimation compared to that without any mutual information estimation (\emph{i.e.} 0.75 AUC). This is due to the improvement of discriminative ability for the latent feature learned by the mutual information estimators. We also observe that global mutual information estimation (\emph{i.e.} global loss Eq. (\ref{mut6})) achieves the best performance when using only one mutual information estimation loss. Note that training our model with two mutual information estimation losses achieves comparable results to the full three losses (see the 6th, 7th and 8th column in Tab. \ref{losscomposition}). Overall, three mutual information estimation losses indeed help to improve the performance of our model.

We further evaluate the mutual information estimation can help the model to obtain more discriminative feature representation by visualization. To this end, we randomly select 500 in-class samples (frog class) and 500 out-of-class samples from the dataset for testing \footnote{We run data selection 5 times and obtain a similar visualization result. For simplicity, we select one of them for visualization.}.	As illustrated in Fig. \ref{latent}, two global latent features are generated by our framework with/without mutual information estimation on CIFAR-10 by using t-SNE \cite{vanDerMaaten2008}. The normal samples (indicated by blue dots in Fig. \ref{latent} left) are significantly separated from novelty samples (indicated by yellow dots in Fig. \ref{latent} right) by using mutual information estimation, compared to that without this estimator.

	\paragraph{The effect to remove  $z_A$ or $z$.}
 In local and global mutual information estimation, the concatenation of feature maps and latent vector are used as the inputs of their estimators during training. The discriminative feature representation is learned by distinguishing the concatenated features all from positive samples and part from negative samples. To evaluate the effectiveness of the concatenated features, we remove the information of latent matrix $z_A$ in the local estimator or latent vector $z$ in the global estimator. In the first part of Tab. \ref{ab}, the (first 3 rows) combination without any removal in our estimators achieves the highest AUC, compared to the removals of $z_A$ or $z$. 
	
	\paragraph{The effect of  Laplacian pyramid loss.}
 As shown in the second part of Tab. \ref{ab}, the framework without any reconstruction loss results in the worst average AUC score of 0.53. We further compare the Laplacian pyramid loss with the MSE loss. As presented in the second part of Tab. \ref{ab}, the performance of Laplacian pyramid loss significantly outperforms the MSE loss (\emph{i.e.} 0.843 AUC \emph{vs.} 0.810 AUC). To explain, Laplacian pyramid loss is a perception-level error, which is more effective for novelty detection than MSE loss.

	\paragraph{The effect with/without PCA initialization.}
We compared our PCA initialization with normal distribution initialization and contrastive prior. Contrastive prior only uses the output $z$ of the contrastive network at the 100th epoch training as initialization. As shown in the third part of Tab. \ref{ab}, PCA initialization achieves the best performance, compared to other initialization methods. This is because PCA encodes the information of the entire training data, which alleviates the problem of mode dropping.

		\begin{table}[t]
			\scriptsize
			\centering 
			\begin{tabular}{c|c|c|c}
				\hline 
				&\textbf{MNIST} &\textbf{COIL}&\textbf{fMNIST}\\
				\hline 
				\textbf{ALOCC DR  ('18) }&0.88&0.809&0.753\\
				
				\textbf{ALOCC D  ('18)}&0.82&0.686&0.601\\
				
				\textbf{DCAE  ('14) }&0.899&0.949&0.908\\
				
				\textbf{GPND  ('18)}&0.932&0.968&0.901 \\
				
				\textbf{Rot  ('18)}&0.933& 0.970 &0.935 \\
				
				\textbf{OCGAN  ('19)}&0.977&0.995&0.924 \\
				
				\textbf{DualGAN  ('20)}&0.985&\textbf{1.0}&0.995 \\
				\hline
				\textbf{Ours}&\textbf{0.986}&\textbf{1.0}&\textbf{0.999} \\
				\hline 
			\end{tabular}	
			\caption{ Mean AUC results of the different methods on MNIST, COIL and fMNIST.}
			\label{Protocol1}
		\end{table}

	\subsection{Comparison with State-of-the-art Methods}
	
	\subsubsection {Novelty Detection} We first evaluate the effectiveness of our method on CIFAR-10. As shown in Tab. \ref{cifar10},  our method achieves the highest mean AUC of 0.843, compared to other SOTA methods. We also found that the self-supervised learning methods based on the pretext tasks (e.g., RotNet \cite{komodakis2018unsupervised}, Geometric \cite{hendrycks2019using}) achieves higher performance, compared with GAN-based methods (e.g., OCGAN \cite{perera2019ocgan} and DualGAN \cite{ijcai2020-106}).	For the individual class, our method also shows the best results, except the class car, dog, horse and truck. The semantic information of these classes has high related to the object shape in CIFAR-10. Since the RotNet focuses on the semantic information of object shape, it achieves high performance in these classes. However, it lack capturing other semantic feature(\emph{e.g.} object color and object texture).

			\begin{table}[t]
				\tiny
				\centering
				\begin{tabular}{c|c|c|c|c|c}
					\hline 
					\textbf{Scenarios}&
					
					\begin{tabular}[c]{@{}l@{}}\textbf{RotNet}\\\textbf{ \hspace{0.5em}('18)}\end{tabular}
					&
					\begin{tabular} [c]{@{}l@{}}\textbf{Geometric}\\\textbf{\hspace{1.2em}('19)}\end{tabular}
					&
					\begin{tabular}[c]{@{}l@{}}\textbf{WaveNet}\\\textbf{\hspace{1em}('19)}\end{tabular}
					&
					\begin{tabular}[c]{@{}l@{}}\textbf{DualGAN}\\\textbf{\hspace{1.1em}('20)}\end{tabular}
					
					&\textbf{
						Ours
					} \\
					\hline 
					\textbf{Beach}&0.508&0.522&0.72&0.82&\textbf{0.86}\\
					
					\textbf{Bus}&0.562&0.585&0.83&\textbf{0.96}&\textbf{0.96}\\
					
					\textbf{restaurant}&0.507&0.560&0.76&\textbf{0.80}&0.78\\
					
					\textbf{Car}&0.606&0.664&0.82&\textbf{0.99}&\textbf{0.99}\\
					
					\textbf{City center}&0.510&0.532&0.82&0.89&\textbf{0.90}\\
					
					\textbf{Forest path}&0.515&0.530&0.72&0.78&\textbf{0.80}\\
					
					\textbf{Grocery store}&0.511&0.530&0.77&0.90&\textbf{0.95}\\
					
					\textbf{Home}&0.504&0.511&0.69&\textbf{0.90}&0.68\\
					
					\textbf{Library}&0.514&0.531&0.67&0.89&\textbf{0.97}\\
					
					\textbf{Metro station}&0.512&0.533&0.79&0.89&\textbf{0.93}\\
					
					\textbf{Office}&0.508&0.523&0.78&0.87&\textbf{0.94}\\
					
					\textbf{Park}&0.520&0.550&0.80&0.95&\textbf{0.99}\\
					
					\textbf{Residential area}&0.512&0.527&0.78&0.78& \textbf{0.81}\\
					
					\textbf{Train}&0.522&0.567&0.84&0.92&\textbf{0.95}\\
					
					\textbf{Tram}&0.568&0.577&0.87&\textbf{0.97}&\textbf{0.97}\\
					\hline 
				\end{tabular}
				\caption{AUC scores for all methods on DCASE dataset with 15 scenarios.}
				\label{voice_tab}
			\end{table}

Following \cite{perera2019ocgan}, we also evaluate the performance of novelty detection methods on COIL100, MNIST, fMNIST by using the following evaluation setting: The 80\% of in-class samples are regarded as a normal class for training, while the rest of 20\% of in-class samples is adopt for testing.  Out-of-class test samples have the same number of in-class test samples, which are randomly selected from the test set. As shown in Tab. \ref{Protocol1}, our method achieves the best performance across different datasets, compared to state-of-the-art methods.  For example, in MNIST dataset and fMNIST dataset, we achieve the improvement of average AUC score 0.1\% and  0.4\% respectively.

\subsubsection{Acoustic Anomaly Detection}
\begin{figure}[t]
	\centering
	\includegraphics[width= \linewidth]{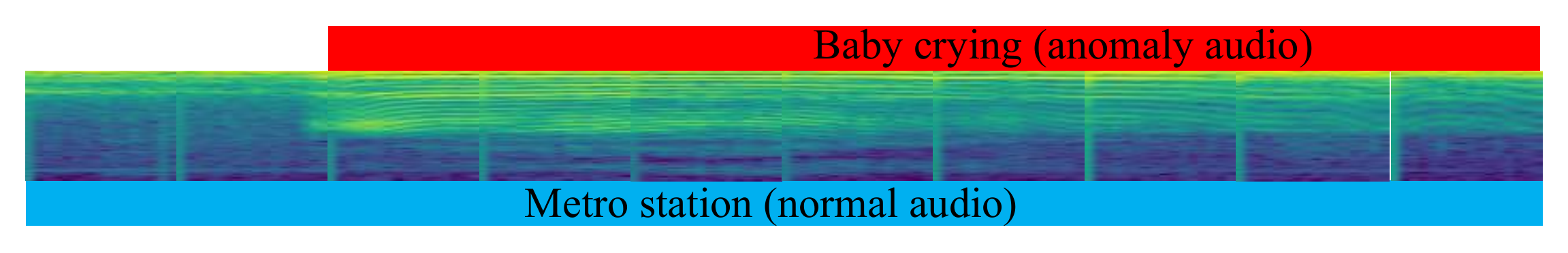}
	\caption{The visualization of test sample: Metro background audio mixed with baby crying.}
	\label{voice}
\end{figure}

The original WaveNet \cite{oord2016wavenet} has been successfully applied into raw audio generation and music synthesis, which benefits from its powerful convolutional autoregressive architecture. Recently, \cite{rushe2019anomaly} has extended its structure for anomaly detection in raw audio. Thus, we also make a comparison with it on the DCASE dataset, which is denoted by WaveNet for convenience. As shown in Tab. \ref{voice_tab}, we present the results of different methods across 15 classes/scenarios. Obviously, our method achieves the best performance in most of the scenarios, except the restaurant and home scenarios, compared to CAE, WaveNet, and DualGAN. For example, our method outperforms the best DualGAN by 8\% AUC on library scenario. Note that DualGAN achieves amazing performance with 0.9 AUC. We conjecture this is due to the randomness of the DualGAN method in the home background. Interestingly, self-supervised learning based methods (\emph{e.g.} RotNet \cite{komodakis2018unsupervised} and Geometric \cite{hendrycks2019using}) shows worse performance on DCASE. This is due to the failure to presentation of object colors and object textures. Actually, as shown in Fig. \ref{voice}, the main discriminative features between normal and novelty audio in spectrogram are from object colors and textures. Rich semantic information of object shapes obtained by these methods cannot separate the normal samples from novelty samples. 

	\subsubsection{Spoofing Face Detection}
	Our method also works for spoofing face detection. Actually, we formulate face anti-spoofing detection as a novelty detection task by only using the normal (live faces) samples for training. We intuitively take the optical flow obtained from consecutive video frames as input, since the data source of spoof face detection is from the video. To this end, we first extract the optical flow using FlowNet2.0 \cite{Ilg2017FlowNet2E} from each video with 30 fps.
	
	\begin{table}[t!]
		\scriptsize
		\begin{tabular}{c|c|c}
			\hline \textbf{Methods}&
			\begin{tabular}[c]{@{}l@{}}\textbf{	Train on CASIA-MFSD }\\\textbf{\&Test on Replay-Attack}\end{tabular}
			& 
			\begin{tabular}[c]{@{}l@{}}\textbf{	Train on Replay-Attack }\\\textbf{\&Test on CASIA-MFSD}\end{tabular}		
			\\ \hline
			\textbf{	LBP ('13)  }          & 0.470 & 0.396                                        \\ 
			\textbf{	LBP-TOP  ('13)     }             & 0.497&0.606   \\ 
			\textbf{	Motion  ('13)      }      &0.502& 0.479                                      \\
			\textbf{	CNN   ('14)    }  &0.485& 0.455                                    \\ 
			\textbf{	Color LBP  ('15)    }     & 0.379&0.354                                        \\
			\textbf{	Color Tex  ('16)  }   & 0.303 & 0.377                                         \\ 
			\textbf{	Auxiliary  ('18)   }   &0.276& 0.284                                     \\ 
			\textbf{	De-Spoof ('18)   }     & 0.285&0.411                                      \\ 
			
			\textbf{	DA  ('18)    }         & 0.274& 0.360                      \\ 
			
			\textbf{	D-texture  ('18)    }         & 0.222 & 0.350                    \\ 
			
			\textbf{	OF Domain  ('18)    }         & 0.301& 0.368                        \\ 
			
			\textbf{	ADA  ('19)    }        & \textbf{0.175} & 0.416                        \\ 
			
			\textbf{	GFA-CNN  ('20)    }         & 0.214& 0.343                         \\ 
			
			\textbf{	DualGAN ('20)    }         & 0.223& \textbf{0.246}                        \\ 
			\hline
			\textbf{	Ours   }     & \textbf{0.175}& 0.308                        \\ \hline
		\end{tabular}
		\caption{Classification performance of the proposed approach in terms of HTER.}
		\label{crossdataset}
	\end{table}
	
	Following \cite{tu2020learning}, the detection models select the training set from one of the training set in CASIA-MFSD and Replay-Attack dataset, and the test set from the other test dataset. The results are summarized in Tab. \ref{crossdataset}. The proposed method achieves the best performance (HTER = 0.175) on the Replay-Attack test set of which includes different types of spoofing attacks. On the other dataset setting, our method achieves a competitive performance (HTER = 0.308) on the testing set of the CASIA-MFSD dataset. Actually, the HTER achieved by our unsupervised method significantly lower than supervision methods except auxiliary method \cite{liu2018learning}. This is probably due to the help of the additional depth information and rPPG signal. Nevertheless, our method achieves better performance by simultaneously evaluating these two tasks on the average HTER. We also found that the generalization of these models trained on the CASIA-MFSD dataset is better than the models trained on the Replay-Attack dataset. We speculate that the detection scenario on the CASIA-MFSD dataset is more complex than the Replay-Attack dataset, which leads to the easy learning of more knowledge during training. 

\begin{figure}[t]
	\centering
	\includegraphics[width= \linewidth]{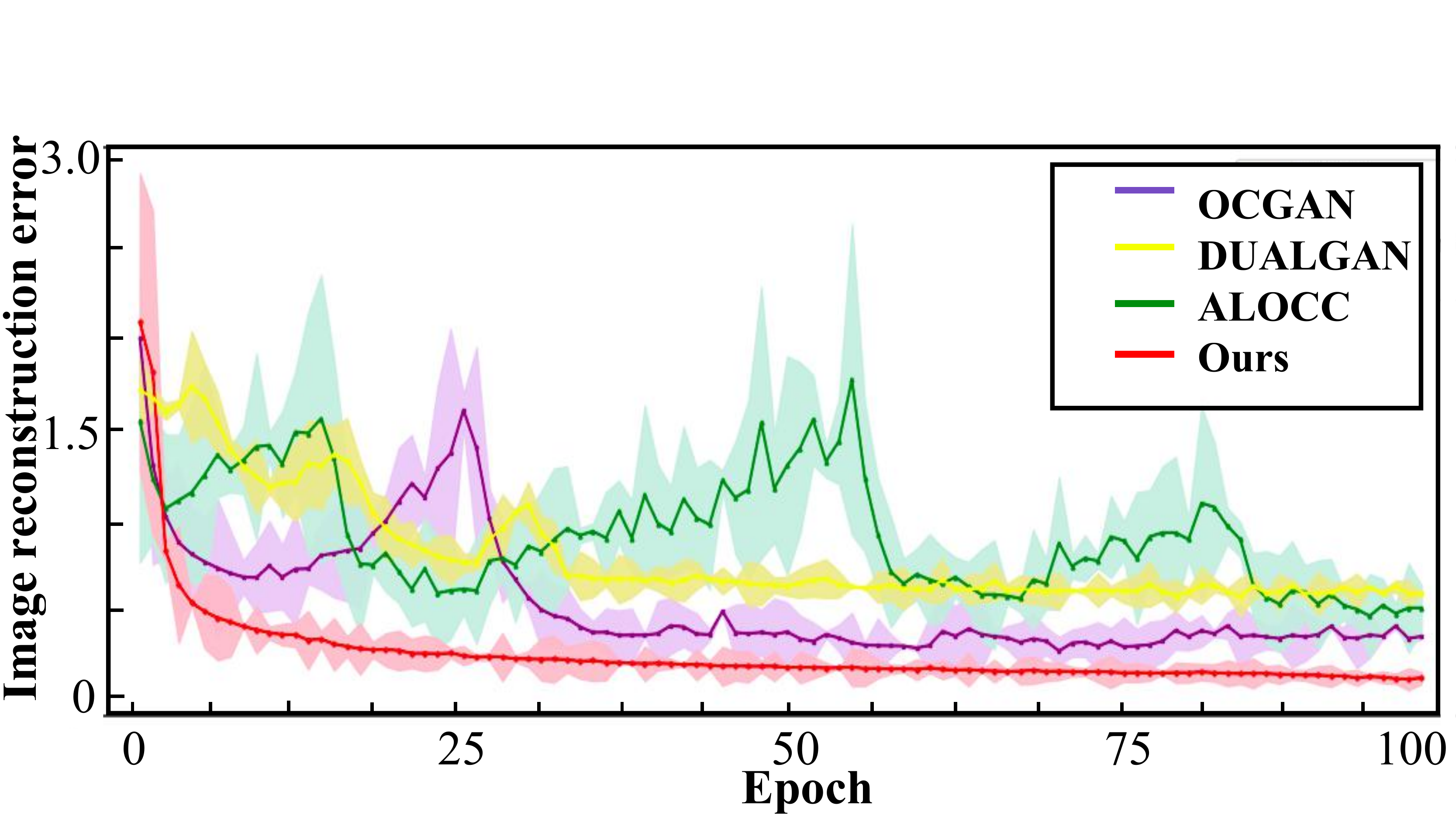}
	\caption{The visualization of training loss comparison between previous methods and proposed method on CIFAR-10.}
	\label{imageloss}
\end{figure} 

\subsection {Stability and Convergence}\label{subsec:Stability} We further evaluate the effectiveness of our method during training. Our decoder-encoder framework achieves more stable training and fast convergence speed in CIFAR-10 dataset, compared to other encoder-decoder frameworks, \emph{e.g.} OCGAN, ALOCC and DualGAN. As shown in Fig. \ref{imageloss}, the proposed decoder-encoder framework tends to be convergent after the 25-th epoch, while the convergent value of other methods is significantly larger (\emph{e.g.} about 40 epochs in OCGAN). In addition, the image reconstruction error is reduced steadily using the decoder-encoder framework, while the significant large fluctuations occur in other methods. This is due to the usage of the mutual information estimator that adversarial optimization is removed in our framework.

\section{Conclusion}
In this paper, we propose a novel decoder-encoder framework for novelty detection. To alleviate mode dropping of GANs, each latent initialized vector is mapped to an image by the PCA initialization in the generative network. To learn a more discriminative latent feature representation, we introduce a contrastive network to learn to compare the different input pairs through mutual information estimation. Specifically, our framework is trained without adversarial optimization, which benefits fast convergence and stable training of our model. We have comprehensively evaluated the performance of our method on a variety of novelty detection tasks over different datasets, which demonstrates the superior performance gains over the state-of-the-art methods.

\section*{Acknowledgments}
\noindent We thank for the support from National Natural Science Foundation of China (61772524, 61902129, 61972157, 61876161, 61701235, 61373077), Shanghai Pujiang Talent Program (19PJ1403100), the Science and Technology Commission of Pudong (NO. PKJ2018-Y46) and Shanghai Jiaotong University Translational Medicine Cross Foundation (ZH2018ZDA25), Shanghai Sailing Program (21YF1411200), Natural Science Foundation of Shanghai (20ZR1417700), the Fundamental Research Funds for the Central Universities, CAAI-Huawei MindSpore Open Fund.

\bibliography{ijcai21}

\begin{thebibliography}{}

\bibitem[\protect\citeauthoryear{Arora \bgroup \em et al.\egroup
  }{2018}]{arora2017gans}
Sanjeev Arora, Andrej Risteski, and Yi~Zhang.
\newblock Do gans learn the distribution? some theory and empirics.
\newblock In {\em ICLR}, 2018.

\bibitem[\protect\citeauthoryear{Belghazi \bgroup \em et al.\egroup
  }{2018}]{belghazi2018mine}
Mohamed~Ishmael Belghazi, Aristide Baratin, Sai Rajeshwar, Sherjil Ozair,
  Yoshua Bengio, Aaron Courville, and Devon Hjelm.
\newblock Mutual information neural estimation.
\newblock In {\em ICML}, 2018.

\bibitem[\protect\citeauthoryear{Bengio and
  Mari{\'e}thoz}{2004}]{bengio2004statistical}
Samy Bengio and Johnny Mari{\'e}thoz.
\newblock A statistical significance test for person authentication.
\newblock In {\em Odyssey}, 2004.

\bibitem[\protect\citeauthoryear{Chen \bgroup \em et al.\egroup
  }{2020a}]{ijcai2020-106}
Chengwei Chen, Jing Liu, Yuan Xie, Yin~Xiao Ban, Chunyun Wu, Yiqing Tao, and
  Haichuan Song.
\newblock Latent regularized generative dual adversarial network for abnormal
  detection.
\newblock In {\em IJCAI}, 2020.

\bibitem[\protect\citeauthoryear{Chen \bgroup \em et al.\egroup
  }{2020b}]{chen2020simple}
Ting Chen, Simon Kornblith, Mohammad Norouzi, and Geoffrey Hinton.
\newblock A simple framework for contrastive learning of visual
  representations.
\newblock In {\em ICML}, 2020.

\bibitem[\protect\citeauthoryear{Chingovska \bgroup \em et al.\egroup
  }{2012}]{chingovska2012effectiveness}
Ivana Chingovska, Andr{\'e} Anjos, and S{\'e}bastien Marcel.
\newblock On the effectiveness of local binary patterns in face anti-spoofing.
\newblock In {\em BIOSIG}, 2012.

\bibitem[\protect\citeauthoryear{Hendrycks \bgroup \em et al.\egroup
  }{2019}]{hendrycks2019using}
Dan Hendrycks, Mantas Mazeika, Saurav Kadavath, and Dawn Song.
\newblock Using self-supervised learning can improve model robustness and
  uncertainty.
\newblock In {\em NeurIPS}, 2019.

\bibitem[\protect\citeauthoryear{Hjelm \bgroup \em et al.\egroup
  }{2019}]{hjelm2019learning}
R~Devon Hjelm, Alex Fedorov, Samuel Lavoie-Marchildon, Karan Grewal, Phil
  Bachman, Adam Trischler, and Yoshua Bengio.
\newblock Learning deep representations by mutual information estimation and
  maximization.
\newblock In {\em ICLR}, 2019.

\bibitem[\protect\citeauthoryear{Ilg \bgroup \em et al.\egroup
  }{2017}]{Ilg2017FlowNet2E}
Eddy Ilg, Nikolaus Mayer, Tonmoy Saikia, Margret Keuper, Alexey Dosovitskiy,
  and Thomas Brox.
\newblock Flownet 2.0: Evolution of optical flow estimation with deep networks.
\newblock In {\em CVPR}, 2017.

\bibitem[\protect\citeauthoryear{Ji \bgroup \em et al.\egroup
  }{2019}]{ji2019invariant}
Xu~Ji, Jo{\~a}o~F Henriques, and Andrea Vedaldi.
\newblock Invariant information clustering for unsupervised image
  classification and segmentation.
\newblock In {\em ICCV}, 2019.

\bibitem[\protect\citeauthoryear{Komodakis and
  Gidaris}{2018}]{komodakis2018unsupervised}
Nikos Komodakis and Spyros Gidaris.
\newblock Unsupervised representation learning by predicting image rotations.
\newblock In {\em ICLR}, 2018.

\bibitem[\protect\citeauthoryear{Krizhevsky and
  Hinton}{2009}]{krizhevsky2009cifar}
Alex. Krizhevsky and Geoffrey. Hinton.
\newblock Learning multiple layers of features from tiny images.
\newblock {\em Master's thesis, Department of Computer Science, University of
  Toronto}, 2009.

\bibitem[\protect\citeauthoryear{Lecun and Bottou}{1998}]{1998Gradient}
Yann Lecun and Leon Bottou.
\newblock Gradient-based learning applied to document recognition.
\newblock {\em Proceedings of the IEEE}, 86(11):2278--2324, 1998.

\bibitem[\protect\citeauthoryear{Lim \bgroup \em et al.\egroup
  }{2018}]{lim2018doping}
Swee~Kiat Lim, Yi~Loo, Ngoc-Trung Tran, Ngai-Man Cheung, Gemma Roig, and Yuval
  Elovici.
\newblock Doping: Generative data augmentation for unsupervised anomaly
  detection with gan.
\newblock In {\em ICDM}, 2018.

\bibitem[\protect\citeauthoryear{Ling and Okada}{2006}]{ling2006diffusion}
Haibin Ling and Kazunori Okada.
\newblock Diffusion distance for histogram comparison.
\newblock In {\em CVPR}, 2006.

\bibitem[\protect\citeauthoryear{Liu \bgroup \em et al.\egroup
  }{2018}]{liu2018learning}
Yaojie Liu, Amin Jourabloo, and Xiaoming Liu.
\newblock Learning deep models for face anti-spoofing: Binary or auxiliary
  supervision.
\newblock In {\em CVPR}, 2018.

\bibitem[\protect\citeauthoryear{Liu \bgroup \em et al.\egroup
  }{2020}]{liu2020self}
Xiao Liu, Fanjin Zhang, Zhenyu Hou, Zhaoyu Wang, Li~Mian, Jing Zhang, and Jie
  Tang.
\newblock Self-supervised learning: Generative or contrastive.
\newblock {\em arXiv preprint arXiv:2006.08218}, 1(2), 2020.

\bibitem[\protect\citeauthoryear{Marchi \bgroup \em et al.\egroup
  }{2017}]{marchi2017deep}
Erik Marchi, Fabio Vesperini, Stefano Squartini, and Bj{\"o}rn Schuller.
\newblock Deep recurrent neural network-based autoencoders for acoustic novelty
  detection.
\newblock {\em Computational intelligence and neuroscience}, 2017, 2017.

\bibitem[\protect\citeauthoryear{Mesaros \bgroup \em et al.\egroup
  }{2017}]{DCASE2017challenge}
Annamaria Mesaros, Toni Heittola, Aleksandr Diment, Benjamin Elizalde, Ankit
  Shah, Emmanuel Vincent, Bhiksha Raj, and Tuomas Virtanen.
\newblock Dcase 2017 challenge setup: Tasks, datasets and baseline system.
\newblock In {\em DCASE 2017-Workshop on Detection and Classification of
  Acoustic Scenes and Events}, 2017.

\bibitem[\protect\citeauthoryear{Nene \bgroup \em et al.\egroup
  }{1996}]{Nene1996}
Samer~A. Nene, Shree~K. Nayar, and Hiroshi Murase.
\newblock Columbia object image library (coil-20).
\newblock Technical Report CUCS-005-96, Department of Computer Science,
  Columbia University, February 1996.

\bibitem[\protect\citeauthoryear{Nowozin \bgroup \em et al.\egroup
  }{2016}]{nowozin2016f}
Sebastian Nowozin, Botond Cseke, and Ryota Tomioka.
\newblock f-gan: Training generative neural samplers using variational
  divergence minimization.
\newblock In {\em NeurIPS}, 2016.

\bibitem[\protect\citeauthoryear{Oord \bgroup \em et al.\egroup
  }{2016}]{oord2016wavenet}
Aaron van~den Oord, Sander Dieleman, Heiga Zen, Karen Simonyan, Oriol Vinyals,
  Alex Graves, Nal Kalchbrenner, Andrew Senior, and Koray Kavukcuoglu.
\newblock Wavenet: A generative model for raw audio.
\newblock {\em arXiv preprint arXiv:1609.03499}, 2016.

\bibitem[\protect\citeauthoryear{Paszke \bgroup \em et al.\egroup
  }{2019}]{paszke2017automatic}
Adam Paszke, Sam Gross, Francisco Massa, Adam Lerer, James Bradbury, Gregory
  Chanan, Trevor Killeen, Zeming Lin, Natalia Gimelshein, Luca Antiga, Alban
  Desmaison, Andreas Kopf, Edward Yang, Zachary DeVito, Martin Raison, Alykhan
  Tejani, Sasank Chilamkurthy, Benoit Steiner, Lu~Fang, Junjie Bai, and Soumith
  Chintala.
\newblock Pytorch: An imperative style, high-performance deep learning library.
\newblock In {\em NeurIPS}, 2019.

\bibitem[\protect\citeauthoryear{Perera \bgroup \em et al.\egroup
  }{2019}]{perera2019ocgan}
Pramuditha Perera, Ramesh Nallapati, and Bing Xiang.
\newblock Ocgan: One-class novelty detection using gans with constrained latent
  representations.
\newblock In {\em CVPR}, 2019.

\bibitem[\protect\citeauthoryear{Radford \bgroup \em et al.\egroup
  }{2016}]{radford2015unsupervised}
Alec Radford, Luke Metz, and Soumith Chintala.
\newblock Unsupervised representation learning with deep convolutional
  generative adversarial networks.
\newblock In {\em ICLR}, 2016.

\bibitem[\protect\citeauthoryear{Rushe and Mac~Namee}{2019}]{rushe2019anomaly}
Ellen Rushe and Brian Mac~Namee.
\newblock Anomaly detection in raw audio using deep autoregressive networks.
\newblock In {\em ICASSP}, 2019.

\bibitem[\protect\citeauthoryear{Sabokrou \bgroup \em et al.\egroup
  }{2018}]{sabokrou2018adversarially}
Mohammad Sabokrou, Mohammad Khalooei, Mahmood Fathy, and Ehsan Adeli.
\newblock Adversarially learned one-class classifier for novelty detection.
\newblock In {\em CVPR}, 2018.

\bibitem[\protect\citeauthoryear{Sinha \bgroup \em et al.\egroup
  }{2021}]{sinha2021negative}
Abhishek Sinha, Kumar Ayush, Jiaming Song, Burak Uzkent, Hongxia Jin, and
  Stefano Ermon.
\newblock Negative data augmentation.
\newblock {\em arXiv preprint arXiv:2102.05113}, 2021.

\bibitem[\protect\citeauthoryear{Tack \bgroup \em et al.\egroup
  }{2020}]{tack2020csi}
Jihoon Tack, Sangwoo Mo, Jongheon Jeong, and Jinwoo Shin.
\newblock Csi: Novelty detection via contrastive learning on distributionally
  shifted instances.
\newblock In {\em NeurIPS}, 2020.

\bibitem[\protect\citeauthoryear{Tu \bgroup \em et al.\egroup
  }{2020}]{tu2020learning}
Xiaoguang Tu, Zheng Ma, Jian Zhao, Guodong Du, Mei Xie, and Jiashi Feng.
\newblock Learning generalizable and identity-discriminative representations
  for face anti-spoofing.
\newblock {\em ACM Transactions on Intelligent Systems and Technology (TIST)},
  11(5):1--19, 2020.

\bibitem[\protect\citeauthoryear{van~der Maaten and
  Hinton}{2008}]{vanDerMaaten2008}
Laurens van~der Maaten and Geoffrey Hinton.
\newblock Visualizing data using {t-SNE}.
\newblock {\em Journal of Machine Learning Research}, 9:2579--2605, 2008.

\bibitem[\protect\citeauthoryear{Vincent \bgroup \em et al.\egroup
  }{2010}]{vincent2010stacked}
Pascal Vincent, Hugo Larochelle, Isabelle Lajoie, Yoshua Bengio, and
  Pierre-Antoine Manzagol.
\newblock Stacked denoising autoencoders: Learning useful representations in a
  deep network with a local denoising criterion.
\newblock {\em Journal of machine learning research}, 11(Dec):3371--3408, 2010.

\bibitem[\protect\citeauthoryear{Xiao \bgroup \em et al.\egroup
  }{2017}]{xiao2017/online}
Han Xiao, Kashif Rasul, and Roland Vollgraf.
\newblock Fashion-mnist: a novel image dataset for benchmarking machine
  learning algorithms.
\newblock {\em arXiv preprint arXiv:1708.07747}, 2017.

\bibitem[\protect\citeauthoryear{Zhang \bgroup \em et al.\egroup
  }{2012}]{zhang2012face}
Zhiwei Zhang, Junjie Yan, Sifei Liu, Zhen Lei, Dong Yi, and Stan~Z Li.
\newblock A face antispoofing database with diverse attacks.
\newblock In {\em ICB}, 2012.

\bibitem[\protect\citeauthoryear{Zhao \bgroup \em et al.\egroup
  }{2017}]{zhao2016energy}
Junbo~Jake Zhao, Micha{\"{e}}l Mathieu, and Yann LeCun.
\newblock Energy-based generative adversarial networks.
\newblock In {\em ICLR}, 2017.

\end{thebibliography}
\bibliographystyle{named}

\end{document}